%% file: elsarticle-template.tex
\documentclass[review]{elsarticle}

\usepackage{lineno,hyperref}
\modulolinenumbers[5]

\usepackage{times}
\usepackage{epsfig}
\usepackage{graphicx} 
\usepackage{amsmath}
\usepackage{amssymb}
\usepackage{subfigure}
\usepackage{multirow}
\usepackage{xcolor}
\usepackage{adjustbox}
\usepackage{grffile}

\journal{Pattern Recognition}

\begin{document}

\begin{frontmatter}

\title{AFINet: Attentive Feature Integration Networks for Image Classification}


\author[address1]{Xinglin Pan}
\author[address2]{Jing Xu}
\author[address2]{Yu Pan}
\author[address1]{liangjian Wen}
\author[address3]{WenXiang Lin}
\author[address4]{Kun Bai}
\author[address5,address2]{Zenglin Xu\corref{cor1}}
\cortext[cor1]{Corresponding author}
\ead{xuzenglin@hit.edu.cn}

\address[address1]{University of Electronic Science and Technology of China, Chengdu, China}
\address[address2]{School of Science and Technology, Harbin Institute of Technology, Shenzhen, China}
\address[address3]{Beijing Institute of Technology, Beijing, China}
\address[address4]{Cloud and Smart Industries Group, Tencent, China}
\address[address5]{Artificial Intelligence Center, Peng Cheng Lab, Shenzhen, China}




\begin{abstract}
\input{paper_abstract}
\end{abstract}

\begin{keyword}
CNN \sep Attention \sep Image Classification\sep Feature Integration
\MSC[2010] 68T45
\end{keyword}

\end{frontmatter}

\linenumbers

\input{paper_intro}

\input{paper_related}
\input{paper_model}
\input{paper_exper}

\section{Conclusion}
In this paper, we proposed AFI modules that can adaptively select features for transferring and improve the representational power of networks. The output feature of AFI modules is aggregated by the channel-wise soft attention over a series of features. The structure of AFI modules is simple and can be used directly in existing state-of-the-art residual-like networks easily. Experimental results show the effectiveness of AFI-Networks, which achieves competitive performance on multiple datasets. Compared with baseline models, AFI counterparts can achieve better performance with lower computational complexity. We believe that the AFI modules are broadly applicable across various computer vision tasks, e.g., object detection, instance segmentation and semantic segmentation.

\bibliography{egbib}

\end{document}

%% file: paper_abstract.tex
Convolutional Neural Networks (CNNs) have achieved tremendous success in a number of learning tasks including image classification.
Recent advanced models in CNNs, such as ResNets, mainly focus on the skip connection to avoid gradient vanishing.
DenseNet designs suggest creating additional bypasses to transfer features as an alternative strategy in network design.
In this paper, we design Attentive Feature Integration (AFI) modules, which are widely applicable to most recent network architectures, leading to new architectures named AFI-Nets.
AFI-Nets explicitly model the correlations among different levels of features and selectively transfer features with a little overhead.
AFI-ResNet-152 obtains a 1.24\% relative improvement on the ImageNet dataset while decreases the FLOPs by about 10\% and the number of parameters by about 9.2\% compared to ResNet-152.

%% file: paper_intro.tex
\section{Introduction}
Convolutional Neural Networks(CNNs) have achieved remarkable successes in various computer vision tasks, e.g., image classification, semantic segmentation, object detection~\cite{lecun1995convolutional, gatys2016image,wang2017residual, jiang2013salient}, etc. A major drive to such successes is from the evolutionary design of network architectures, and such representative examples include AlexNet~\cite{krizhevsky2012imagenet}, VGG-Net~\cite{simonyan2014very}, GoogleNet~\cite{szegedy2015going}, ResNet~\cite{he2016deep}, and  DenseNet~\cite{huang2017densely}. Notably, the skip connection introduced in ResNet has become a fundamental design strategy as an effective solution to the gradient vanishing problem, especially for very deep networks. And this strategy has been widely adopted in a variety of architectures, including ShuffleNet~\cite{ma2018shufflenet}, ResNeXt~\cite{xie2017aggregated}, and Inception-ResNet~\cite{szegedy2017inception}, etc. Besides, as an alternative design strategy, DenseNet~\cite{huang2017densely} suggests constructing extra bypasses to transfer previous layers of features for future reuse.


Constructing extra bypasses leads to many advantages in architecture designs, e.g., implicit deep supervision and diversified depths~\cite{huang2017densely,lee2019energy}. 
Furthermore, preserved features, especially low-level features, are beneficial to overcome overfitting (which often leads to small training errors but large testing errors). As evidenced by a study of the class selectivity indices~\cite{morcos2018importance}, the generalization gap (i.e., the difference between the training error and the testing error) of features is prone to increase with depths. Hence, high-level features are more useful to reduce the training error while low-level ones are in favor of closing the generalization gap. These observations suggest that aggregation of features at multiple levels is vital to the designs of architectures.

Despite the advantages of the current bypass design for classification, there are several issues to be addressed: (1) 
Some low-level features (e.g., edges of the background) may be irrelevant to classification and thus impair the learning performance; (2) 
The concatenation of all previous features involves a quadratic scale of memory usage which is inhibited in scenarios with limited computational and storage resources;
(3)
The correlation between features crossing layers is hard to model for the convolution operators. 


\begin{figure*}[t]
\centering
\includegraphics[width=.9\textwidth]{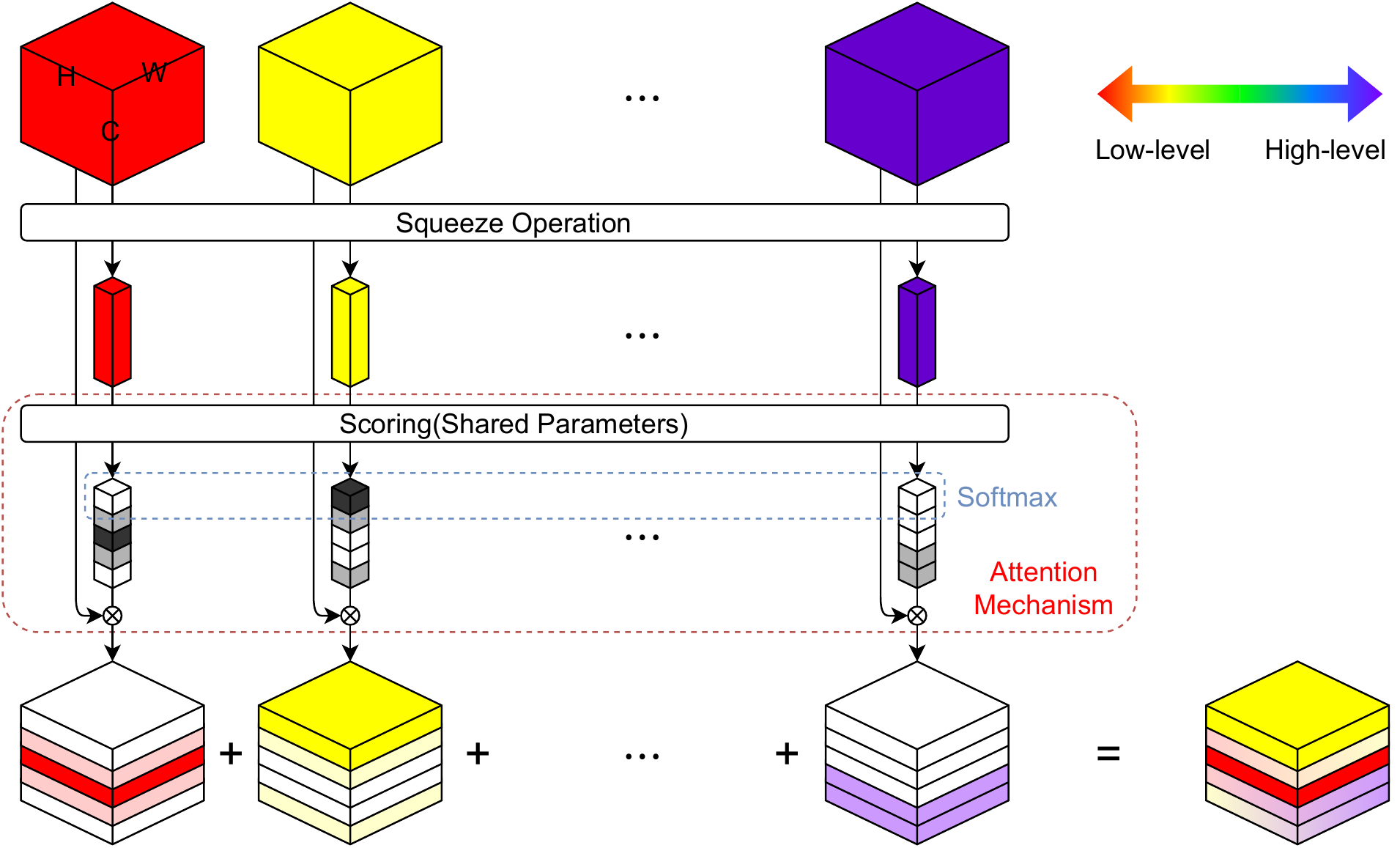}
\caption{The AFI module can automatically extract important low-level features for the high-level features. Through two light-weight operations, i.e., the squeeze operation and the shared attention mechanism,  every feature is re-calibrated along each channel dimension. Finally, the resulting feature is supplied to  later layers.}
\label{fig:fu_data}
\end{figure*}

\begin{figure*}[h!tb]
\centering
\subfigure[Input]{
\begin{minipage}[t]{0.19\linewidth}
\includegraphics[width=1.\textwidth]{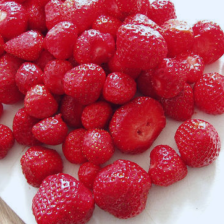}
\end{minipage}%
}%
\subfigure[Conv1-1]{
\begin{minipage}[t]{0.19\linewidth}
\includegraphics[width=1.\textwidth]{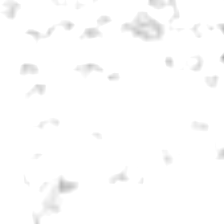}
\end{minipage}%
}%
\subfigure[Conv1-2]{
\begin{minipage}[t]{0.19\linewidth}
\includegraphics[width=1.\textwidth]{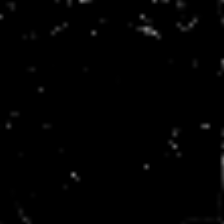}
\end{minipage}%
}%
\subfigure[Conv1-3]{
\begin{minipage}[t]{0.19\linewidth}
\includegraphics[width=1.\textwidth]{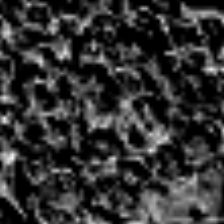}
\end{minipage}%
}%
\subfigure[AFI Output]{
\begin{minipage}[t]{0.19\linewidth}
\includegraphics[width=1.\textwidth]{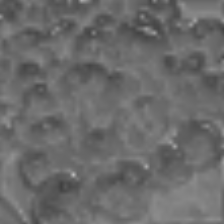}
\end{minipage}%
}%
\caption{The subfigure (a) shows an image from the Imagenet dataset. The subfigure (b), (c) and (d) show features from conv1 \textit{stage} in AFI-ResNet-50. All of these extract edge information. The subfigure (e) depicts the output of an AFI module whose inputs are the subfigure (b), (c), and (d). The subfigure (e) extracts the main edge information of the three subfigures.}
\label{fig:vis}
\end{figure*}

To address these issues, we propose a lightweight and selective feature integration scheme for most residual-like networks, leading to the Attentive Feature Integration (AFI) module, as illustrated in Figure~\ref{fig:fu_data}.
Firstly,  each of the input feature maps (i.e., raw features as shown in Figure~\ref{fig:vis}) with $C$ channels,  is squeezed into a vector by a squeeze operation that captures information from a large spatial extent. Thus, the global context is embedded in the vectors. 
Secondly, the vectors are scored and normalized sequentially via the attention mechanism (which consists of the shared scoring function and the channel-by-channel softmax function) in order to re-calibrate features. At last, we obtain the resulting feature via a summation of re-calibrated features, where each channel can be viewed as a convex combination over the raw features. In short, AFI modules rate the importance of features adaptively and have access to model correlations between distant layers. 

AFI modules are not limited to a special backbone network. Instead, they can be easily plugged to various backbones, leading to various architectures, namely AFI-Nets. To interpret the reason for no limit to a special backbone network, different backbone networks are uniformly viewed as instances of ordinary differential equations(ODEs)~\cite{chang2017multi, chen2018neural,wang2019learning}. The application of AFI modules is very similar to the linear multistep method~\cite{ascher1998computer} apply to solve ODEs. (see Section \ref{ODE} for more about this.)




\begin{figure*}[t]
\centering
     \includegraphics[width=0.8\textwidth]{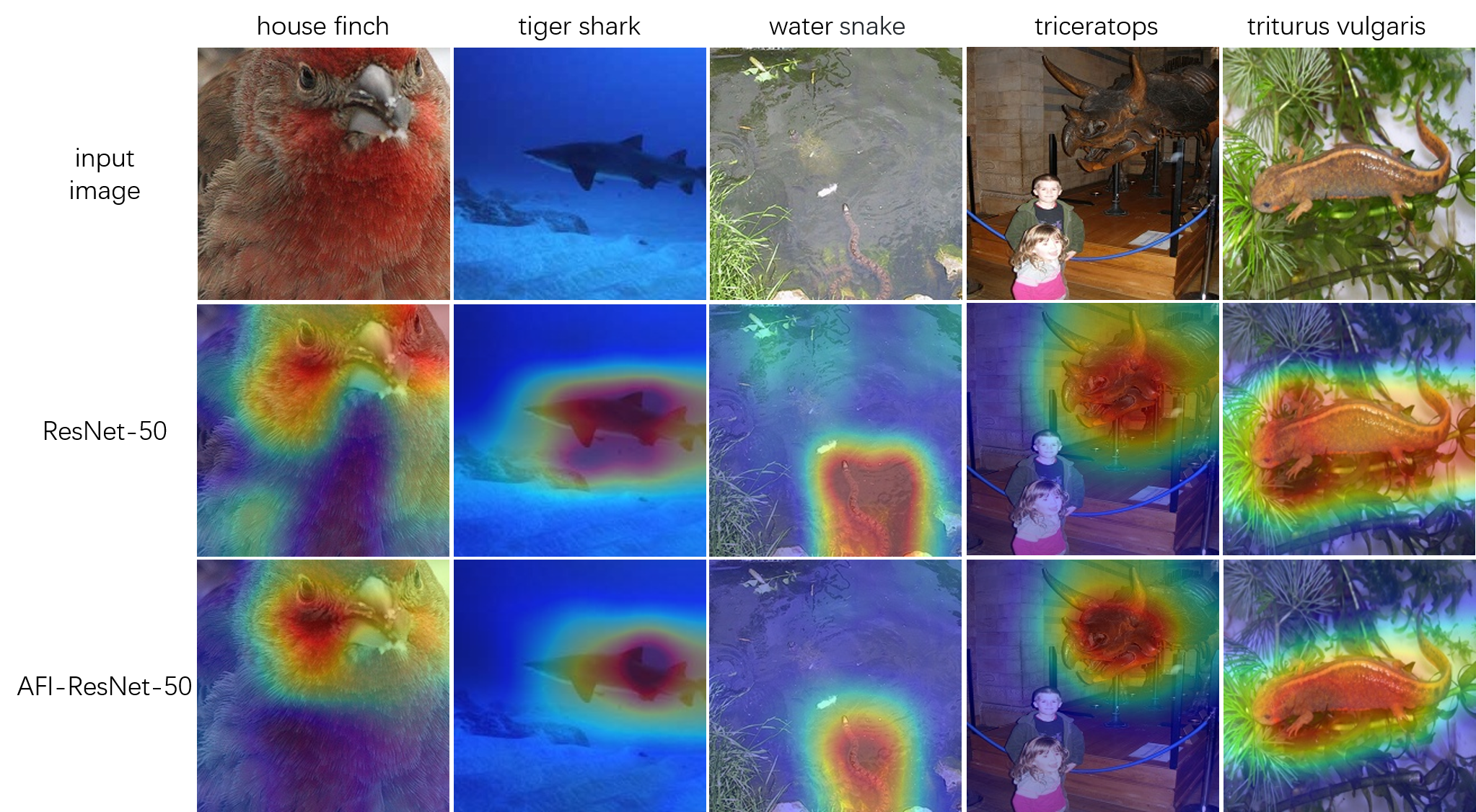}
      \caption{Illustration of the impact of the AFI module. All original images come from the ImageNet dataset. Heatmaps generated by Grad-CAM ~\cite{selvaraju2017grad} illustrates which areas the network pays more attention to. Compared to the vanilla ResNet-50, the area that AFI-ResNet-50 paid most attention to is much smaller.}
      \label{fig:attention}
\end{figure*}

In addition to the easy plug-and-go property, AFI modules also enjoy many advantages such as efficient utilizing of low-level features, and having lower overheads and higher accuracy. 
To illustrate the efficiency, we perform a series of indirect experiments. We firstly visual class activation mapping(CAM) as depicted in Figure \ref{fig:attention}. The activation area of AFI-ResNet-50 is much precise than ResNet-50. Then, the average of the class selectivity index~\cite{morcos2018importance} of features at different layers is lower than backbones networks, which argues that features AFI-Nets learnt are more common. At last, we obtain a 10.8\% average relative improvement on various extremely easy to overfit datasets. The great performance of these experiments is usually attributed to the efficient utilization of low-level features.

\begin{figure*}[htbp]
\centering
     \includegraphics[width=0.8\textwidth]{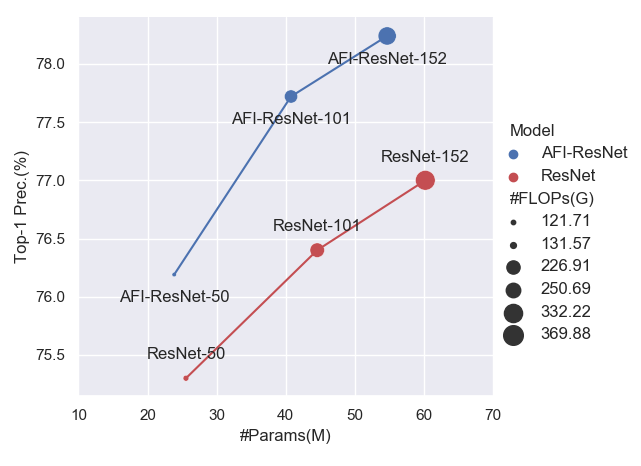}
      \caption{Illustration of the experiment results of AFI-ResNets. With the same number of layers, AFI-ResNets have fewer parameters, fewer FLOPs, and more accuracy. The FLOPs are calculated by assuming the batch size of 32.}
      \label{fig:result}
\end{figure*}

AFI-Nets are trained on ImageNet with SGDM optimizers and regular data augmentations. As shown in Figure \ref{fig:result}, AFI-ResNets are applicable with different depths. With the same number of layers, AFI-ResNet has fewer parameters, fewer FLOPs, and more accuracy. Among AFI-ResNets, AFI-ResNet-152 increases the Top-1 accuracy rate in ImageNet by 1.24\% compared to ResNet-152 while decreases the FLOPs by about 10\% and the number of parameters by about 9.2\%.

Our major contributions can be summarized as follows:
\begin{itemize}
\item We propose lightweight AFI modules to selectively transfer features. From the perspective and inspiration of ODEs, AFI-Nets can be derived by applying AFI modules into residual-like networks.
\item To avoid the difficulty of directly evaluating the efficiency in processing low-level features, we perform three indirect analytical experiments to verify AFI modules take more advantage of preserved low-level features. 
\item Experimental results show that our AFI-Nets significantly improve the representational power of the network.
\end{itemize}

%% file: paper_related.tex
\section{Related Work}
\textbf{Modern CNN Architectures.} Deep convolutional neural networks(CNNs) have dominated image classification since the AlexNet \cite{krizhevsky2012imagenet} and VGG-Net \cite{simonyan2014very} are proposed. After that, substantial efforts have been made to improve the efficiency of CNNs. The modular design strategy in GoogleNet \cite{szegedy2015going} simplifies the network architecture and the multi-path structure in each module shows a great success. ResNet introduces the shortcut connection alleviating the difficulty in deep network training~\cite{he2016deep}. DenseNet \cite{huang2017densely} densely connects all preceding layers to take full advantage of preceding feature maps. Based on those fundamental architectures, some advanced variants(e.g., ResNeSt~\cite{zhang2020resnest}) have been proposed and have achieved impressive performance in many computer vision tasks.

\textbf{Exploitation of Low-Level Features.}
The motivation to preserve low-level features is ample. As shown in Yosinski et al.~\cite{yosinski2014transferable}, some shallow layer filters obtain similar effects as Gabor filters and color blobs. Such low-level features appear to be unspecific to particular datasets or tasks, and generalize well to other datasets or tasks. 
To maximize feature utilization,  DenseNet~\cite{huang2017densely} demonstrates that dense concatenating all the features in frontier layers can effectively alleviate the difficulty of training and improves the network performance with an increased calculation overhead. VoVNet~\cite{lee2019energy}, VoVNetV2~\cite{lee2019centermask} overcome the inefficiency of dense connection by concatenating all features only once in the last feature map and achieve the state-of-the-art performance in instance segmentation. Dual Path Network \cite{chen2017dual} and Mixed link Network \cite{wang2018mixed} try to transfer lowel-level feature based on ResNet~\cite{he2016deep}. However, few works focus on constructing bypass to transfer low-level features based on other residual-like networks.

\textbf{Attention Mechanisms.}
The benefits of attention mechanism have been demonstrated across a range of tasks. Squeeze-and-Excitation block \cite{hu2018squeeze} highly appreciates attention mechanism and thus well improves the accuracy of varied CNNs. They use global average-pooled features to exploit the inter-channel relationship and to compute the channel-wise attention. Besides, there are several other researches to utilize the attention mechanism and improve the results of CNNs in various vision tasks. CBAM~\cite{woo2018cbam} further adds the spatial attention to the SE module and results in better plug-and-play modules.
Soft mask branches to refine the feature maps by adding attention knowledge proposed by \cite{wang2017residual}. Non-local Neural Networks \cite{wang2018non} proposes non-local module to integrate the global attention information. Libra R-CNN \cite{pang2019libra} designs the balanced feature pyramid which refines the semantic feature from multi-level features. BASNet \cite{qin2019basnet} pays more attention to the boundary of the mask by the boundary-aware loss function. However, few literature focuses on the mix of attention mechanism and transferring low-level features. The proposed AFI module thus aims to improve transferring based on attention mechanisms.

%% file: paper_model.tex
\section{Our Model}


In this section, we first introduce Attentive Feature Integration (AFI) modules and compare them with previous works. Next, we describe the ways of implementing AFI modules for residual-like networks with two examples. Specifically, we integrate AFI modules with backbone architectures, such as ResNet~\cite{he2016deep} and MobileNetV2~\cite{sandler2018mobilenetv2}, to build new network architectures, i.e., AFI-ResNet and AFI-MobileNetV2, respectively. At last, inspired by the view of residual-like networks as instances of ordinary differential equations~(ODEs), we point out that our modules is similar to Linear Multistep Method~(LMM) in both motivation and formal.



\subsection{Attentive Feature Integration Modules}


AFI module aims to model the correlation among features at different levels and then produce a compact but comprehensive feature as an output. AFI module is composed of two operations: a squeeze operation and an attention mechanism. For convenient description, we denote $N$ different-level features as $\boldsymbol{X}^{(i)} \in \mathbb{R}^{H \times W \times C}, i\in\{1,2,...,N\}$.

\textbf{Squeeze Operation.} $F_{\text{sq}}(\cdot)$ represents the squeeze function(e.g., global average pooling), which gathers contextual long-range feature interactions, embedding global context into a vector descriptor. By shrinking $\boldsymbol{X}^{(i)}$ on its spatial dimensions $H \times W$, the channel-wise statistic $\boldsymbol{z}^{(i)}\in\mathbb{R}^{C}$ is generated, where $C$ is the number of channels.

\textbf{Attention Mechanism.} The attention mechanism is a selective aggregation of information so that the resulting features are easily exploited for specific tasks. There are two steps in the attention mechanism: score and normalization. The score step is based on a shared scoring function $F_{\text{sc}}(\cdot)$, which is applied into vector descriptors to produce an embedding of importance $\boldsymbol{s}^{(i)}\in\mathbb{R}^{C}$. $F_{\text{sc}}(\cdot)$ is formulated by two transformation matrices around the activation function
\begin{equation}
\boldsymbol{s}^{(i)}=F_{\text{sc}}\left(\boldsymbol{z}^{(i)}, \boldsymbol{W}\right)=\boldsymbol{W}_2\text{ReLU}\left(\boldsymbol{W}_1\boldsymbol{z}^{(i)}\right),
\end{equation}
where the vector $\boldsymbol{s}^{(i)}$ is parameterized by forming a bottleneck layer with a weight $\boldsymbol{W}_1 \in \mathbb{R}^{\frac{C}{r} \times C}$, a dimensionality-increasing layer with a weight $\boldsymbol{W}_2 \in \mathbb{R}^{C \times \frac{C}{r}}$ and the reduction ratio $r$. Next, a matrix $\boldsymbol{S} = \left[\boldsymbol{s}^{(1)}, \boldsymbol{s}^{(2)}, \cdots, \boldsymbol{s}^{(N)}\right] \in \mathbb{R}^{N \times C}$,  is obtained by concatenating $\{s^{(i)}\}_{i=1}^{N}$. Then, a matrix $\widetilde{\boldsymbol{S}} \in \mathbb{R}^{N \times C}$ is formed by normalizing $\boldsymbol{S}$ along the $N$ dimension using a Softmax function.

In summary, $\text{AFI}: \left\{\boldsymbol{X}^{(i)}\right\}_{i=1}^N \rightarrow \boldsymbol{R}$ is formulated by

 
\begin{align}
&\boldsymbol{z}^{(i)} = F_{\text{sq}}\left(\boldsymbol{X}^{(i)}\right),\quad
\boldsymbol{s}^{(i)} = F_{\text{sc}}\left(\boldsymbol{z}^{(i)}, \boldsymbol{W}\right), \quad
\boldsymbol{S} = \left[\boldsymbol{s}^{(1)}, \boldsymbol{s}^{(2)}, \cdots,\boldsymbol{s}^{(N)}\right], \nonumber\\ & \widetilde{\boldsymbol{S}}_{i,j} = \frac{\exp \left(\boldsymbol{S}_{i,j}\right)}{\sum_{k=1}^{N}{\exp \left(\boldsymbol{S}_{k,j}\right)}} , \quad \boldsymbol{R}_{i,j,k}= \sum_{l=1}^{N}{\widetilde{\boldsymbol{S}}_{l,k} \boldsymbol{X}_{i,j,k}^{(l)}}. 
\end{align}
where the resulting feature $\boldsymbol{R}\in \mathbb{R}^{H \times W \times C}$.

\textbf{Remark on the connection to previous works.}
Recently, a series of literature attempts to incorporate the attention mechanism to improve the performance of CNNs. One of the most popular computational units is the Squeeze-and-Excitation(SE) module~\cite{hu2018squeeze}. Compared to SE module, our AFI module focus on explicitly modeling the feature integration instead of channel-wise selection. Moreover, compared to DenseNet with self-attention modules like CAPR-DenseNet~\cite{zhang2019channel}, our module can be applied to deeper and larger networks, while avoiding quadratic complexity of memory usage and running time by substituting the shared attention mechanism for the independent exciter. 
Compared to the SKNet~\cite{li2019selective} that selects the efficient kernel size, our AFI module can utilize different level(e.g., positional and semantic) information. Compare to the Mixed Link Network~\cite{wang2018mixed}, our module can be applied to not only ResNet but also other residual-like networks.


\subsection{AFI Modules for Residual-Like Networks}
This subsection clarifies how to apply AFI module to residual-like networks and derive our AFI-Nets. In order to describe our method more clearly, we follow the definition of \textit{stage} and \textit{building blocks} in Hu et al.~\cite{hu2018genet}. More specifically, a \textit{stage} consists of several \textit{building blocks} with features of the same shape stacking sequentially.

\begin{figure*}[h!tb]
\centering
\subfigure[AFI-ResNet \textit{building block}]{
\begin{minipage}[t]{0.44\linewidth}
\includegraphics[width=.8\textwidth]{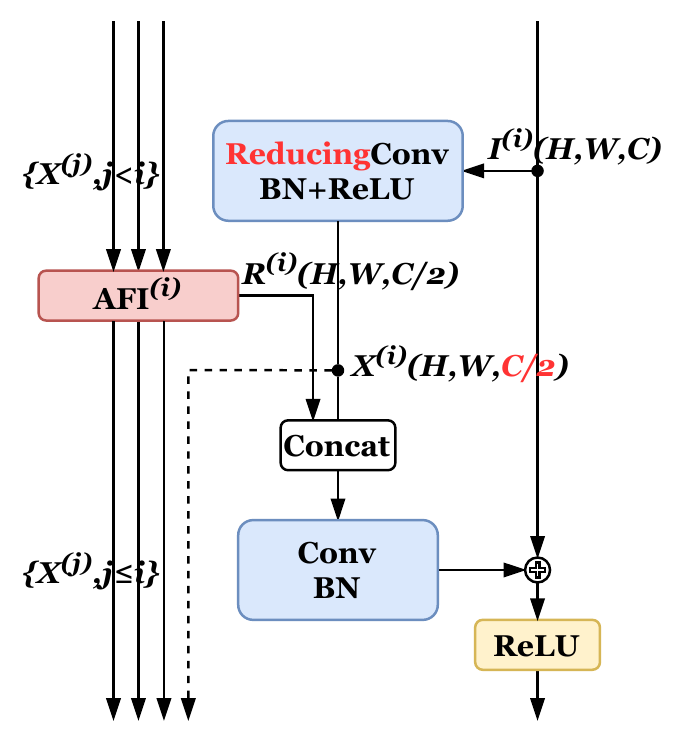}
\label{fig:resnet}
\end{minipage}%
}%
\subfigure[AFI-MobileNet \textit{building block}]{
\begin{minipage}[t]{0.55\linewidth}
\includegraphics[width=.8\textwidth]{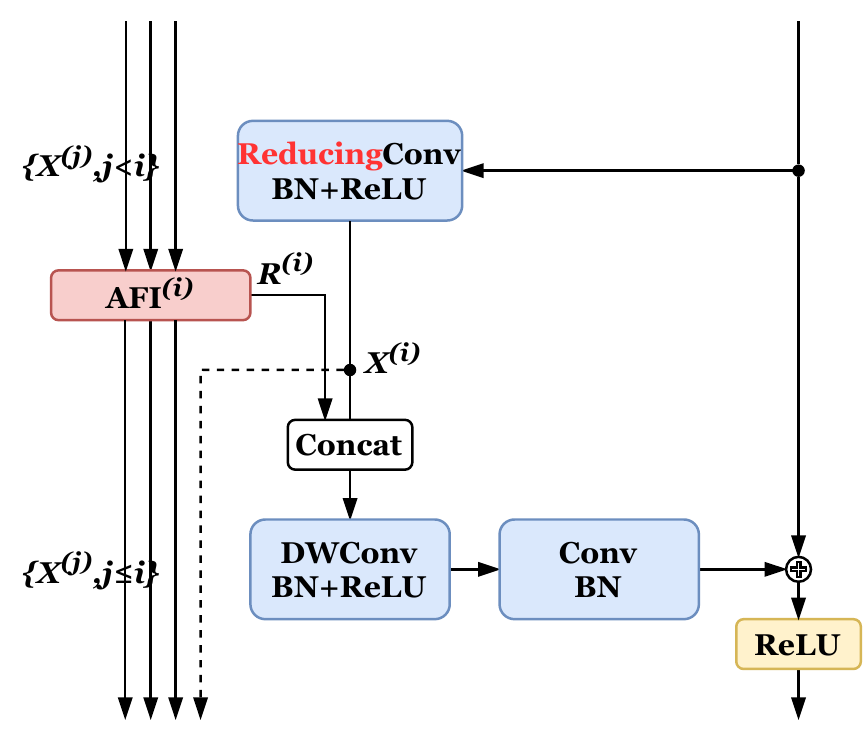}
\label{fig:mobilenet}
\end{minipage}%
}%

\centering
\caption{The architecture of AFI-MobileNet \textit{building block} is shown in Figure \ref{fig:mobilenet}. The architecture of AFI-ResNet \textit{building block} is shown in Figure \ref{fig:resnet}. The blue and yellow parts indicate the original backbone while the red parts indicate our revisions. The position denoted by $\boldsymbol{I}^{(i)}$ is the input of the $i\text{-th}$ \textit{building block} of an original ResNet.}
\label{fig:twoexample}
\end{figure*}

As shown in Figure \ref{fig:twoexample}, we half the output of the first convolution in $i\text{-th}$ building blocks and denote it as $\boldsymbol{X}^{(i)}$. For the $i\text{-th}$ \textit{building block} of each \textit{stage}, $\text{AFI}^{(i)}$ denotes the AFI module corresponding to that \textit{building block}.  $\{\boldsymbol{X}^{(1)}, \boldsymbol{X}^{(2)}, \cdots, \boldsymbol{X}^{(i-1)}\}$ in previous \textit{building blocks} are the inputs of $\text{AFI}^{(i)}$. We concatenate the output $\boldsymbol{R}^{(i)}$ of  $\text{AFI}^{(i)}$ with $\boldsymbol{X}^{(i)}$ and then throw the concatenated result to the subsequent convolution layer. We name our networks as AFI-Nets.

Here, we would like to describe applications of our module in residual-like networks with an instance AFI-ResNet. AFI-ResNet \textit{building block} is shown in Figure \ref{fig:resnet}. We first review the architecture of ResNet \textit{building block}~\cite{he2016deep}. For the $i\text{-th}$ building block of ResNet
\begin{equation}
\boldsymbol{I}^{(i+1)} = \text{ReLU} \left( \boldsymbol{I}^{(i)} + \text{BN}\left(  \text{ReLU} \left( \text{BN} \left(\boldsymbol{I}^{(i)} \ast \boldsymbol{K}^{(i)}_1 \right) \right) \ast \boldsymbol{K}^{(i)}_2 \right)   \right) ,
\end{equation}
where $\ast$ denotes convolution operator, $\text{BN}\left(\cdot\right)$ denotes batch normalization, $\text{ReLU}(\cdot)$ is a rectified linear activation function. The input feature $\boldsymbol{I}^{(i)}$ is shown in Figure \ref{fig:resnet} and the kernels $\boldsymbol{K}^{(i)}_1, \boldsymbol{K}^{(i)}_2 \in \mathbb{R}^{k\times k\times C\times C}$ 
are learnable filters where the kernel size $k=3$ and the input channel and the output channel $C\in\{16, 32, 64\}$.

In AFI-ResNet, we replace $\boldsymbol{K}^{(i)}_1$ with a shrinking kernel $\widetilde{\boldsymbol{K}}^{(i)}_1 \in \mathbb{R}^{k \times k \times C \times \frac{C}{2}}$ to reduce parameters. For the $i\text{-th}$ \textit{building block} of AFI-ResNet
\begin{align}
&\boldsymbol{X}^{(i)} =  \text{ReLU} \left( \text{BN} \left( \boldsymbol{I}^{(i)} \ast \widetilde{\boldsymbol{K}}^{(i)}_1 \right) \right)  , \quad \boldsymbol{R}^{(i)} = \text{AFI}^{(i)} \left( \boldsymbol{X}^{(1)}, \boldsymbol{X}^{(2)}, ..., \boldsymbol{X}^{(i-1)} \right) \nonumber\\ & \boldsymbol{I}^{(i+1)} = \text{ReLU} \left (\boldsymbol{I}^{(i)} + \text{BN} \left([\boldsymbol{X}^{(i)}, \boldsymbol{R}^{(i)}] \ast \boldsymbol{K}^{(i)}_2 \right) \right). 
\label{equ:concat}
\end{align}

Besides, we also provide an instance AFI-MobileNetV2 as shown in \ref{fig:mobilenet}. We concatenate the output of the AFI module with the output feature of the first convolutional layer. 
As many building blocks (e.g., bottleneck \textit{building blocks}~\cite{he2016deep}) use an 1$\times$1 convolution at the front of building blocks to reduce channels, our AFI modules can neutralize the impairment of bottleneck convolution. It is elaborated in Section \ref{overhead} that the parameters and FLOPs of AFI-Nets are usually smaller than the vanilla.

\subsection{The Versatility of AFI modules}
\label{ODE}
In this section, we will explain the versatility of AFI modules from the view of numerical analysis. Residual-like networks can be identified as instances of ordinary differential equations (ODEs), behaving like the forward Euler method with an initial value $y^{(1)}$~\cite{chang2017multi, chen2018neural,wang2019learning}. We firstly review the forward Euler method as background.
Formally,
\begin{align}
y^{(i)} - y^{(i-1)} &= \left(x^{(i)} - x^{(i - 1)}\right)\frac{d y}{d x}+\left(x^{(i)} - x^{(i - 1)}\right)^{2}\frac{\frac{d^2 y}{d x^2}}{2 !}+\cdots \nonumber \\
y^{(i)} - y^{(i - 1)} &\approx \left(x^{(i)} - x^{(i - 1)}\right) \cdot f\left(x^{(i)}, y^{(i)}\right)\ ,
\label{equ:eular}
\end{align}
where $i\in\left\{2,3, ..., N\right\}$ and $f(x, y)$ denotes $\frac{d y}{d x}$. When satisfying $x^{(i)} - x^{(i - 1)} = 1$, Equation~\eqref{equ:eular} can be simplified as
\begin{equation}
y^{(i)}  =  y^{(i - 1)} + f\left(x^{(i)}, y^{(i)}\right).
\end{equation}
Euler method is a traditional first-order solution to the ODE. Therefore, this method can cause huge error in predicting the next value when ignoring higher order $\left\{\frac{d^i y}{d x^i}\right\}_{i=2}^\infty$. Linear Multi-step Method (LMM)~\cite{ascher1998computer} reuses the information in the previous steps by linear combination to fit them:
\begin{equation}
y^{(i)}  =  y^{(i - 1)} + \sum_{j=1}^{i} \alpha^{(i)}_j f\left(x^{(j)}, y^{(j)}\right),
\label{equ:scalar_lmm}
\end{equation}
where $\{\alpha^{(i)}_j\}_{j=1}^{i}$ are coefficients of merging previous steps.

In residual-like networks, lower-level features play similar roles to the previous steps, thus it is also necessary to take these features into consideration. Thus, based on the consideration, we design AFI-Nets, which is the generalization of LMM. 
Specifically, we first replace the scalars in Equation~\eqref{equ:scalar_lmm} with the features in neural networks, namely
\begin{equation}
\boldsymbol{I}^{(i+1)} = \boldsymbol{I}^{(i)} + g^{(i)}\left(\sum_{j=1}^{i} \boldsymbol{\alpha}^{(i)}_j f\left(\boldsymbol{I}^{(j)}\right)\right)\ ,
\label{equ:lmm}
\end{equation}
where $f\left(\boldsymbol{I}^{(j)}\right)$ is a learnable function that acts similar to estimating $\left.\frac{dy}{dx}\right|_{x=x^{(j)}}$, and $g^{(i)}$ is a function to match the shape of $\boldsymbol{I}^{(i)}$ and the estimated residual. For instance, in AFI-ResNet, $f\left(\boldsymbol{I}^{(j)}\right)$ is taken to be $\text{ReLU}\left(\text{BN}\left(\boldsymbol{I}^{(j)} \ast \widetilde{\boldsymbol{K}}_1^{(j)} \right)\right) $ with a kernel $\widetilde{\boldsymbol{K}}_1^{(j)}$, $g^{(i)}\left(\cdot\right)$ is taken to be $\text{BN}\left(\cdot\ast \hat{\boldsymbol{K}}_2^{(i)} \right)$ with a kernel $\hat{\boldsymbol{K}}_2^{(i)}$, $\left\{\boldsymbol{\alpha}^{(i)}_j, j < i \right\}$ is selected by $\text{AFI}^{(i)}$, and $\boldsymbol{\alpha}^{(i)}_i = \boldsymbol{1}$. Equation~(\ref{equ:lmm}) can be simplified as:
\begin{align}
&g^{(i)}\left(\sum_{j=1}^{i} \boldsymbol{\alpha}^{(i)}_j f\left(\boldsymbol{I}^{(j)}\right)\right) \nonumber \\
&g^{(i)}\left(\sum_{j=1}^{i - 1} \boldsymbol{\alpha}^{(i)}_j \boldsymbol{X}^{(j)} +\boldsymbol{X}^{(i)}\right) \nonumber \\
=&  \text{BN}\left( \left( \boldsymbol{R}^{(i)} + \boldsymbol{X}^{(i)} \right) \ast \hat{\boldsymbol{K}}_2^{(i)}\right) \nonumber \\
=&  \text{BN}\left( \boldsymbol{X}^{(i)} \ast \hat{\boldsymbol{K}}_2^{(i)} + \boldsymbol{R}^{(i)} \ast \hat{\boldsymbol{K}}_2^{(i)} \right)\ .
\label{equ:add}
\end{align}
Compared with $\boldsymbol{K}_2^{(i)}$ in Equation~(\ref{equ:concat}), $\hat{\boldsymbol{K}}_2^{(i)}$ in Equation~(\ref{equ:add})  is a special case of $\boldsymbol{K}_2^{(i)}$ when $\boldsymbol{X}^{(i)}$ and $\boldsymbol{R}^{(i)}$ is applied to convolution with the same kernel. Other AFI-Nets can be expressed by adopting different $f$ and $g$. 



%% file: paper_exper.tex
\section{Experiment}
We firstly analyse overheads of our modules. Then, before presenting results on real-world datasets, we first illustrate the role of our AFI module by several experiments. If not otherwise specified, the reduction ratio $r$ is 4. We show the ablation study to better understand the settings of the AFI modules in last.

\subsection{Overheads of AFI modules}
\label{overhead}

\begin{table}[h!tb]
    \centering
    \caption{The architecture details of AFI-ResNet-(6N+2) for CIFAR dataset. The operations and feature shapes are listed inside the brackets and the number of stacked blocks is shown outside. num\_class depends on the categories.}
    \begin{tabular}{c|c|c}
    \hline
    Name & Output Size & AFI-ResNet-(6N+2) \\
    \hline
    \hline
      Conv$_0$ & 32$\times$32 & 3$\times$3, 16 \\
    \hline
      Conv$_1$ & 32$\times$32 &$
                             \left[
                             \begin{array}{cc}
                             \multirow{2}{*}{AFI +}&3\times 3, 8\\
                               ~ & 3\times 3, 16\\
                              \end{array}
                              \right]\times N
                            $ \\
\hline
      Conv$_2$ & 16$\times$16 & $
                             \left[
                             \begin{array}{cc}
                             \multirow{2}{*}{AFI +} & 
                                3\times 3, 16\\
                               ~ & 3\times 3, 32\\
                              \end{array}
                              \right]\times N
                            $ \\
\hline
      Conv$_3$ & 8$\times$8 &  $
                             \left[
                             \begin{array}{cc}
                             \multirow{2}{*}{AFI +} & 
                                3\times 3, 32\\
                               ~ & 3\times 3, 64\\
                              \end{array}
                              \right]\times N
                            $ \\
\hline
Output &num\_class& AP, FC, Softmax \\
\hline
    \end{tabular}
    \label{tab:detailofresnet}
\end{table}

Additional AFI modules may raise a problem whether parameters and FLOPs are increased. To clarify it, we compare AFI-Nets with the vanilla about parameters and FLOPs. Table~\ref{tab:detailofresnet} shows AFI-ResNet architecture for instance. By setting $N=5$ and $N=18$ separately, AFI-ResNet-32 and AFI-ResNet-110 are acquired. In a $(N+1)\text{-block}$ \textit{stage}, the number of additional parameters is $2 \times (N-1) \times C \times \frac{C}{r}$ and the times of extra multiplication in the scoring function is given by:
\begin{equation}
\sum_{i=3}^{N+1}{2 \times (i - 1) \times \frac{C}{r}\times C}=  \frac{C^2}{r} \times (N+2)(N-1)\ .
\end{equation}
The reason for enumerating $i$ starting from 3 is that, there is no input to $\text{AFI}^{(1)}$ and there is only one input to $\text{AFI}^{(2)}$. The input feature can skip $\text{AFI}^{(2)}$ and is transferred immediately. 

\begin{table*}[h!tb]
    \centering
    \caption{Comparison between our AFI module and convolution to highlight the smartness of our block. No matter in which blocks, with AFI module, the number of parameters and FLOPs are always smaller compared with original ResNet. The bold shows the least parameters and FLOPs. The FLOPs are calculated by assuming the batch size of 32.}
    \begin{tabular}{c|c|c|c|c|c}
    \hline
    Stage & $H\&W$ & $C$ & Network & \#Params & \#FLOPs  \\
    \hline
        \multirow{2}{*}{Stage 1} & \multirow{2}{*}{32} & \multirow{2}{*}{16} & ResNet-32 & 23.36K & 765.46M \\
        \cline{4-6}
        ~ & ~ & ~ & AFI-ResNet-32 & \textbf{18.82K} & \textbf{612.38M} \\
    \hline
        \multirow{2}{*}{Stage 2} & \multirow{2}{*}{16} & \multirow{2}{*}{32} & ResNet-32 & 88.77K & 727.19M \\
        \cline{4-6}
        ~ & ~ & ~ & AFI-ResNet-32 & \textbf{70.72K} & \textbf{575.20M} \\
    \hline
        \multirow{2}{*}{Stage 3} & \multirow{2}{*}{8} & \multirow{2}{*}{64} & ResNet-32 & 353.66K & 724.30M \\
        \cline{4-6}
        ~ & ~ & ~ & AFI-ResNet-32 & \textbf{281.73K} & \textbf{573.02M} \\
    \hline
    \end{tabular}
    \label{tab:my_label}
\end{table*}

It is difficult to compare \textit{building blocks} with AFI modules to the original because the FLOPs of the AFI module are related to the \textit{building block} number $N$ while the FLOPs of convolution operation is related to the size of images. For easier comparison, Table \ref{tab:my_label} shows the differences in parameters and FLOPs between ResNet and AFI-ResNet. All results is calculated by the tool\footnote{https://github.com/Lyken17/pytorch-OpCounter}. In each \textit{stage}, the number of parameters and FLOPs in the AFI module are always smaller than the vanilla ones. Moreover, in most cases, our module assists the original block with decreasing parameters. For example, when the batch size is 32, the ResNet-110 costs 8.17G FLOPs while the AFI-ResNet-110 requires only 6.23G FLOPs, which corresponds to about 24\% decrease.

\subsection{Explore the Role of AFI Module}


\textbf{Class Activation Visualization.}
To illustrate whether the low-level features are collected by our AFI module, we generate the heatmap by Grad-CAM~\cite{selvaraju2017grad}. As we can see in the Figure \ref{fig:attention}, our network has extracted more details of detected objects, and meanwhile, the heatmap generated by ResNet-50 is too smooth to draw the specific borderline of detected objects. With the help of preserving low-level features, the tiny borderlines are more clearer shown in the figure. Besides, compared to ResNet-50, the area that AFI-ResNet-50 paid most attention to decreases obviously.

\begin{figure*}[h!tb]
\centering
     \includegraphics[width=0.98\textwidth]{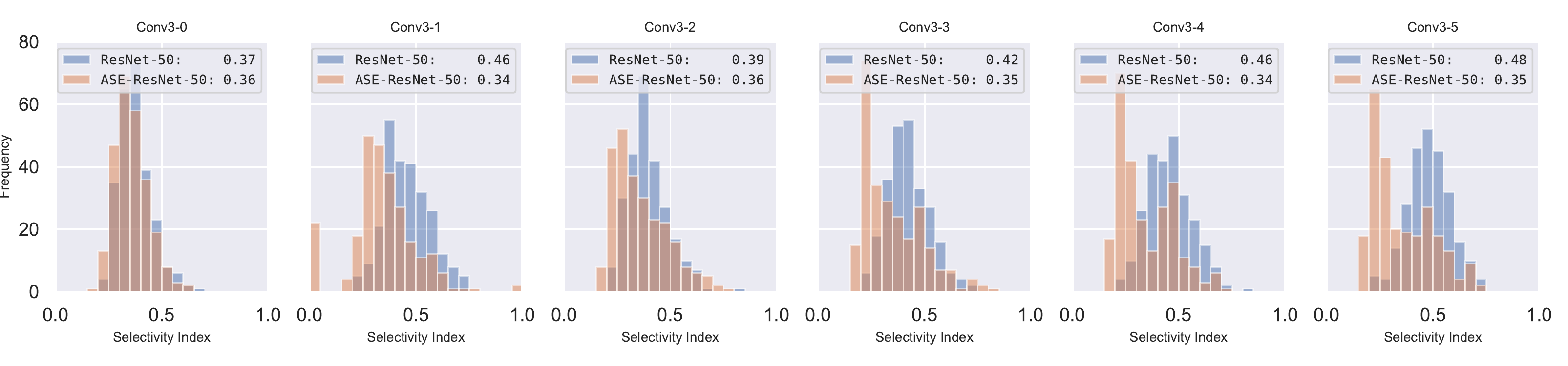}
      \caption{Each figure depicts the class selectivity index distribution for features in both the baseline ResNet-50 and AFI-ResNet-50 various building blocks in the $Conv3$ stage of their architectures. The distributions come from the output of the AFI module and corresponding position of the vanilla in each building block. The mean value is reported after the label.}
      \label{fig:selective}
\end{figure*}

\textbf{Class Selectivity Index.}
Besides, we adopt the class selectivity index metric~\cite{morcos2018importance} to analyze the semantic meaning of features. This metric computes, for each feature map, the difference between the highest class-conditional mean activity and the mean of all remaining class-conditional activities over the testing dataset. The measurement is normalized between zero and one where one indicates that a filter only fires for a single class and zero indicates that the filter produces the class-agnostic value. The less class selectivity index contend to the more generalization of channels in a degree. As shown in Figure \ref{fig:selective}, the AFI-ResNet-50 is able to learn tinier and more generalized features instead of features of a specific class than the vanilla in $Conv3$ \textit{stage}. With the help of generalized features, the generalization gap between the training and the testing dataset will be closed.

\begin{table*}[h!tb]
\caption{The table shows the accuracy rates (\%) of networks on the Office-Home dataset. A:B denotes the task that a model fine-tunes on domain A and tests on domain B.}
    \centering
    \begin{adjustbox}{max width=\textwidth}
    \begin{tabular}{l|c|c|c|c|c|c|c|c|c|c|c|c|c}
    \hline
    Model & Ar:Cl & Ar:Pr & Ar:Rw & Cl:Ar & Cl:Pr & Cl:Rw & Pr:Al & Pr:Cl & Pr:Rw & Rw:Ar & Rw:Cl & Rw:Pr & Avg.\\
    \hline
    \hline
    ResNet-50~\cite{he2016deep} & 34.9&50.0&58.0&37.4&41.9&46.2&38.5&31.2&60.4&53.9&41.2&59.9&46.1 \\
    \hline
    AFI-ResNet-50& 42.0 &64.0& 71.6& 48.2& 56.8& 61.5& 48.5& 36.4& 69.5& 63.7& 45.3& 75.5 & 56.9 \\
    \hline
    \end{tabular}
    \end{adjustbox}
    \label{tab:officeresult}
\end{table*}

\textbf{Performances on a Dataset with the Wide Generalization Gap.}  
We consider the situation where the training dataset is quite different from the testing dataset to verify whether our module is able to bridge the generalization gap. We use another experiment to test whether our module is able to capture the task-agnostic feature which can bridge the generalization gap. We consider the situation where the training dataset is quite different from the test dataset. The better performance indicates that the model has greater generalization ability. 


To satisfy it, we choose the Office-Home dataset~\cite{venkateswara2017deep} which consists of four distinct domains (Artistic images~(Ar), Clip Art~(Cl), Product images~(Pr), and Real-World images~(Rw)). We train the model on ImageNet on the pre-training process, fine-tune on the source domain (A) and then test on the target domain (B). This task is denoted as A:B as shown in Table~\ref{tab:officeresult}. The pre-training details are similar to the training details in Section~(\ref{sec:imagenet}). The epoch of fine-tuning is 20. The learning rate of new full connection layer is $0.01 \times \left(1+10\times \frac{\text{iteration}}{\text{total\_iteration}}\right)^{-0.75}$. The learning rate of the other layers is one-tenth of that of the new full connection layer.

Table~\ref{tab:officeresult} shows the comparison of various domain adaptation task accuracy between ResNet-50 and AFI-ResNet-50. When the generalization gap between the training dataset and the test dataset larger, the advantage of our network that adapts to learn more generalized features is remarked.

The experiment results show that our module can transfer low-level features more efficiently.
In fact, our module provides prior knowledge that a convex combination of features can extract the main information. 
There are some previous methods that assuming additional prior knowledge regularizes the network.
For instance, Mixup~\cite{zhang2017mixup} extends the training distribution by incorporating the prior knowledge that linear interpolations of features should lead to linear interpolations of classified space.

\subsection{Experiments on Real-World Datasets}

\textbf{CIFAR.}
The CIFAR~\cite{krizhevsky2009learning} dataset consists of 60,000 RGB pictures, each with a size of 32$\times$32. 50,000 of them are used as the training set and 10,000 are used for testing. The CIFAR-10 task requires the network to correctly classify the pictures into 10 categories, such as airplanes and automobiles. CIFAR-100 requires the network to classify pictures into 100 categories. We train our network on the training dataset and evaluate it on the test dataset.

By integrating the AFI module with ResNet, ShuffleNetV2, MobileNetV2 and ResNext, we get their AFI counterparts. All the backbone networks have a residual mapping. So we apply the AFI module to the first convolution layer in building blocks to avoid affecting residual mapping and meanwhile neutralize the waste of the bottleneck convolution. 

In this experiment, we set SGD with a momentum of 0.9 and a weight decay of 1e-4. We train the networks with the batch size to 64 for 300 epochs. The learning rate is initialized to 0.1 and divided by 10 at 50\%, 75\% of training process, respectively. Proportion is adopted in \cite{huang2017densely}.
Data augmentation(mirroring/shifting) is used in training. Because MobileNetV2~\cite{sandler2018mobilenetv2}, ShuffleNetV2~\cite{ma2018shufflenet}, and other networks are not designed for the CIFAR dataset, we adopt their variants from Github\footnote{https://github.com/kuangliu/pytorch-cifar}. All results are reproduced and the same experiment settings are adopted for a fair comparison.

\begin{table}[h!tb]
\caption{Accuracy rates (\%) on CIFAR-100 dataset. All results are reproduced by ourselves for a fair comparison. Our network results are bold in the table. The FLOPs are calculated by setting the batch size to 32.}
    \centering
    \begin{tabular}{l|c|c|c}
    \hline
    Model & C100 & \#Params & \#FLOPs \\
    \hline
    \hline
    ResNet-32~\cite{he2016deep} & 71.16 & 472.76K & 2.23G \\
    \hline
    AFI-ResNet-32 & $\textbf{71.09}$ & 378.23K & 1.78G \\
    \hline
    \hline
    ResNet-110~\cite{he2016deep} & 73.73 & 1.74M &8.17G \\
    \hline
    AFI-ResNet-110 & $\textbf{74.03}$ & 1.35M & 6.23G \\
    \hline
    \hline
    ShuffleNetV2~\cite{ma2018shufflenet} & 70.71 & 0.94M & 1.32G \\
    \hline
    AFI-ShuffleNetV2 & $\textbf{72.06}$ & 0.95M & 1.32G \\
    \hline
    \hline
    MobileNetV2~\cite{sandler2018mobilenetv2} & 75.20 & 2.41M & 3.03G \\
    \hline
    AFI-MobileNetV2 & $ \textbf{75.94} $ & 2.25M &2.67G\\
    \hline
    DenseNet-40~\cite{huang2017densely} & 75.58 & 1.06M & 34.48G\\ 
    \hline
    \hline
    ResNext-29(32$\times$4d)~\cite{xie2017aggregated} & 78.44 & 4.87M & 24.95G \\
    \hline
    AFI-ResNext-29(32$\times$4d) & $\textbf{79.37}$ & 4.26M & 21.74G \\
    \hline
    DenseNet-100~\cite{huang2017densely} & 79.80 & 7.09M & 230.119G \\
    \hline
    \end{tabular}
    \label{tab:cifarresult}
\end{table}

Table~\ref{tab:cifarresult} shows the comparison of classification error between the original networks and their corresponding AFI counterparts. As we can see, most of networks work better in classification with assistance of our AFI module. For example, AFI-ResNext29(32$\times$4d) increases the accuracy rate by 0.93\% compared with ResNext29(32$\times$4d)~\cite{xie2017aggregated} on the CIFAR-100 dataset. Besides, as shown in the results, other AFI-Nets also increase the accuracy rates while decrease or remain at least the number of parameters and FLOPs. We additional compare DenseNets~\cite{huang2017densely} with AFI-Nets. When the accuracy of both is comparable, DenseNets~\cite{huang2017densely} require much more FLOPs than AFI-Nets.

\textbf{ImageNet.}
\label{sec:imagenet}
The effect of the AFI module is also evaluated on the ImageNet 2012 dataset~\cite{simonyan2014very} which composes about 1.3 million training images and 50k validation images. Both top-1 and top-5 classification accuracy rates are reported on the validation dataset.

In this experiment, we use SGD with a momentum of 0.9 and a weight decay of 1e-4. We train the networks with batch size 64 for 90 epoch. The initial learning rate is 0.1 * batch\_size / 256  and divided by 10 at 30, 60, and 80 epochs, respectively. 224$\times$224 images serving as the inputs of the network are cropped from the resized raw images or their horizontal flips. Data augmentation in \cite{li2018recurrent} is used in training. We evaluate our model by applying a center-crop with 224$\times$224. 

\begin{table*}[h!tb]
\caption{The table shows the accuracy rates (\%) of networks on the ImageNet validation set. Our results are marked in bold. All results are reproduced for a fair comparison. The FLOPs are calculated by assuming the batch size of 32.}
    
    \centering
    \begin{adjustbox}{max width=\textwidth}
    \begin{tabular}{l|c|c|c|c}
    \hline
    Model & Top-1 Prec. & Top-5 Prec. & \#Params(M) & \#FLOPs(G) \\
    \hline
    \hline
    ResNet-50~\cite{he2016deep} & 75.3 & 92.2 & 25.56 & 131.57 \\
    \hline
    $\textbf{AFI-ResNet-50}$ & 
    \textbf{76.19}$_{(+0.89)}$ &
    \textbf{92.88}$_{(+0.68)}$&
    \textbf{23.85}$_{(-1.71)}$&
    \textbf{121.72}$_{(-9.85)}$\\
    \hline
    \hline
    ResNet-101~\cite{he2016deep} & 76.4 & 92.9 & 44.55 & 250.69 \\
    \hline
    $\textbf{AFI-ResNet-101}$ & 
    \textbf{77.72}$_{(+1.32)}$ &
    \textbf{93.76}$_{(+0.86)}$&
    \textbf{40.75}$_{(-3.80)}$& 
    \textbf{226.91}$_{(-23.78)}$\\
    \hline
    \hline
    ResNet-152~\cite{he2016deep}& 77.0 & 93.3 & 60.19 & 369.88 \\
    \hline
    $\textbf{AFI-ResNet-152}$ & \textbf{78.24}$_{(+1.24)}$ &\textbf{93.98}$_{(+0.68)}$&
    \textbf{54.67}$_{(-5.52)}$ &
    \textbf{332.24}$_{(-37.64)}$ \\
    \hline
    \hline
    MobileNetV2~\cite{sandler2018mobilenetv2}& 66.09 & 87.14 & 3.51 & 10.46 \\
    \hline
    $\textbf{AFI-MobileNetV2}$ &  \textbf{68.24}$_{(+2.15)}$ &\textbf{88.54}$_{(+1.40)}$&
    \textbf{3.63}$_{(+0.12)}$ &
    \textbf{9.36}$_{(-1.10)}$ \\
    \hline
    \hline
    ResNeXt-50(32$\times$4d)~\cite{xie2017aggregated}& 76.08 & 92.92 & 25.03 & 136.30 \\ 
    \hline
    $\textbf{AFI-ResNeXt-50}$ &  \textbf{77.16}$_{(+1.08)}$ &\textbf{93.40}$_{(+0.48)}$&
    \textbf{21.84}$_{(-3.19)}$ &
    \textbf{116.61}$_{(-19.69)}$ \\
    \hline
    \end{tabular}
    \end{adjustbox}
    \label{tab:imagenetresult}
\end{table*}

The accuracy rates of baseline models and our network on the ImageNet validation set are shown in Table \ref{tab:imagenetresult}. With the same backbone architecture, our model always obtains a higher accuracy rate compared with ResNet. For instance, The AFI-ResNet-152 increases the accuracy rate by 1.24\%, decreases the number of parameters by 9.2\%, and meanwhile decreases FLOPs by 10\% compared with ResNet-152.

\begin{table}[h!tb]
    \centering
    \caption{The comparison of the memory usage of AFI-ResNet-50 and ResNet-50 in the training and testing process. The batch size is 64.}
    \begin{tabular}{l|c|c}
    \hline
      Model & ResNet-50 & AFI-ResNet-50 \\
      \hline  
      Training Memory (MiB) & 7447  & \textbf{7001}\\
      \hline
      Inference Memory (MiB) & 3827 & \textbf{2671} \\
      \hline
    \end{tabular}
    \label{tab:memory_cost}
\end{table}

Additionally, the memory usage of our model is smaller than the base model(ResNet-50) in both training and testing process as shown in Table \ref{tab:memory_cost}. In contrast with DenseNet-121~\cite{huang2017densely} which requires an enormous running space for keeping features of the total 24-layer in DenseNet block (3),  our AFI-ResNet-50 maintains features at most 6 layers, therefore the memory usage of our model is much smaller. Besides, reduction of the output of convolution also helps to optimize memory costs during running time. Furthermore, our AFI-Module could release more running space via reducing intermediate gradients in memory with the method proposed by \cite{pleiss2017memory}.

\subsection{Ablation Study}
\textbf{The Position of AFI Module.}
\begin{table*}[h!tb]
    \centering
    \caption{The accuracy rate (\%) comparison of applying our AFI module to different residual stages of ResNet-32 and ResNet-110. Best results are marked in bold.}
    
    \begin{tabular}{c|c|c|c|c}
    \hline
       AFI Stage & ResNet-32 & Stage 1 & Stage 2 & Stage 3  \\
       \hline
        C100 & 71.16 & \textbf{71.38}$_{(+0.22)}$  & 70.63$_{(-0.53)}$ & 70.75$_{(-0.41)}$  \\
        \hline
        \hline
        AFI Stage & ResNet-110 & Stage 1 & Stage 2 & Stage 3  \\
       \hline
        C100 & 73.73 & \textbf{75.03}$_{(+1.30)}$  & 74.13$_{(+0.40)}$ & 74.02$_{(+0.29)}$  \\
        \hline
        
    \end{tabular}
    \label{tab:ablationstudy}
\end{table*}
In this ablation study, we study whether low-level feature maps or high-level feature maps are more hospitable for exploitation by the AFI module. Previous study \cite{yosinski2014transferable} shows that the feature maps learned by the earlier convolutional layers are more general. According to Table \ref{tab:ablationstudy}, by only applying our AFI module to the \textit{stage} 1 of ResNet-32 or ResNet-110, the accuracy rate increases obviously, which means low-level features have more practical impacts.

\begin{table}[h!tb]
    \centering
    \caption{The table shows the accuracy rates (\%) of AFI-ResNet-50 with different $r$ on the ImageNet validation set.}
    \begin{tabular}{c|c|c|c|c}
    \hline
      $r$ & 1 & 2 & 4 & 8 \\
      \hline  
      \hline
      Top-1 Prec. & 76.21 & 76.21 & 76.19 & 76.11\\
      \hline
      \end{tabular}
    \label{tab:r}
\end{table}

\textbf{The Setting of Hyperparameter $r$. }
Tables~\ref{tab:r} shows our AFI module with different reduction ratio $r$.The results show that our AFI module offers a trade-off between improved accuracy and increased model complexity for the real situations. For instance, for a mobile user, the company can adopt AFI-Nets with larger reduction ratio (i.e. $r = 4$) while a computer user can adopt AFI-Nets with smaller reduction ratio (i.e. $r = 2$) to get higher accuracy.

\begin{table*}[h!tb]
\caption{The accuracy rate(\%) of AFI-MobileNetV2 with various of the self-attention model in CIFAR-100.}
\centering
\begin{adjustbox}{max width=\textwidth}
    \begin{tabular}{l|c|c|c|c|c|c}
    \hline
      MobileNetV2  & Baseline & AFI & AFI-SE & AFI-GE & AFI-CBAM & AFI-ECA \\
      \hline  
      C100 & 75.20 & 75.94 & 76.15 & 76.22 & 76.51 & 76.76 \\ 
      \hline
    \end{tabular}
\end{adjustbox}
    \label{tab:c100}
\end{table*}

\textbf{Compatibility with Other Self-attention Models. } Intuitively, the AFI-module is a feature-wise attention mechanism; alternatively, the self-attention module aims to model interdependence between channels explicitly. We further conducted experiments on CIFAR-100 to demonstrate compatibility. In the experiments, we utilize the self-attention modules(SE~\cite{hu2018squeeze}, GE~\cite{hu2018genet}, CBAM~\cite{woo2018cbam}, ECA~\cite{wang2020eca}) to the resulting features. As shown in Table~\ref{tab:c100}, all the AFI-SE, AFI-GE, AFI-CBAM, and AFI-ECA$_{(+0.82\%)}$ models get better results, which indicates that our model is compatible with other self-attention modules.